\documentclass[conference]{IEEEtran}
\IEEEoverridecommandlockouts
\usepackage{cite}

\usepackage{amsmath,amssymb,amsfonts}
\usepackage{algorithmic}
\usepackage{graphicx}
\usepackage{textcomp}
\usepackage{xcolor}
\def\BibTeX{{\rm B\kern-.05em{\sc i\kern-.025em b}\kern-.08em
    T\kern-.1667em\lower.7ex\hbox{E}\kern-.125emX}}

\usepackage[table]{xcolor} 
\usepackage{enumitem}
\usepackage{mathrsfs}%
\usepackage{textcomp}%
\usepackage{manyfoot}%
\usepackage{booktabs}%

\usepackage{listings}%
\usepackage{colortbl}
\usepackage{array}
\usepackage{pgfplots}
\usepackage{adjustbox}
\usepackage{amssymb}
\usepackage{makecell} 

\usepackage[margin=1in]{geometry} 
\usepackage{tikz}
\usetikzlibrary{shapes.multipart, arrows.meta, positioning, calc, decorations.pathreplacing, patterns}

\usepackage{caption}

\usepackage{newunicodechar}
\usepackage{booktabs} 
\usepackage{siunitx} 
\usepackage{multirow} 

\usepackage[linesnumbered,ruled,vlined]{algorithm2e}

\usepackage{bbm}
\usepackage[utf8]{inputenc}
\usepackage[T1]{fontenc}
\usepackage{subcaption}
\usepackage{pgfplots}
\pgfplotsset{compat=1.17}

\usepackage[utf8]{inputenc}
\usepackage[T1]{fontenc}
\usepackage{subcaption}
\usepackage{booktabs}
\usepackage{pgfplots}
\pgfplotsset{compat=1.17}

\usepackage{tikz}
\usepackage{eso-pic}
\usepackage{xcolor}
\definecolor{ijcnnnotice}{gray}{0.55} 

\newcommand{\ijcnnacceptedbanner}{%
  \AddToShipoutPictureFG*{%
    \begin{tikzpicture}[remember picture,overlay]
      \node[anchor=north] at ([yshift=-9mm]current page.north){%
        \parbox{0.92\paperwidth}{\centering
          {\normalfont\rmfamily\footnotesize\color{ijcnnnotice}
            This paper has been accepted for publication at the\\
            IEEE International Joint Conference on Neural Networks (IJCNN), Rome, Italy 2025%
          }%
        }%
      };
    \end{tikzpicture}%
  }%
}

\begin{document}
 \ijcnnacceptedbanner              

\title{BATR-FST: Bi-Level Adaptive Token Refinement for Few-Shot Transformers
}


\author{
\IEEEauthorblockN{1\textsuperscript{st} Mohammed Al-Habib}
\IEEEauthorblockA{
\textit{School of Computer Science and} \\
\textit{Engineering} \\
\textit{Central South University} \\
Changsha, 410083, China \\
alhabib@csu.edu.cn
}
\and
\IEEEauthorblockN{2\textsuperscript{nd} Zuping Zhang\textsuperscript{\textasteriskcentered}}
\IEEEauthorblockA{
\textit{School of Computer Science and} \\
\textit{Engineering} \\
\textit{Central South University} \\
Changsha, 410083, China \\
zpzhang@csu.edu.cn \\
}

\and
\IEEEauthorblockN{3\textsuperscript{rd} Abdulrahman Noman}
\IEEEauthorblockA{
\textit{School of Computer Science and} \\
\textit{Engineering} \\
\textit{Central South University} \\
Changsha, 410083, China \\
abdulrahman@csu.edu.cn
}
}

\maketitle

\begin{abstract}
Vision Transformers (ViTs) have shown significant promise in computer vision applications. However, their performance in few-shot learning is limited by challenges in refining token-level interactions, struggling with limited training data, and developing a strong inductive bias. Existing methods often depend on inflexible token matching or basic similarity measures, which limit the effective incorporation of global context and localized feature refinement. To address these challenges, we propose Bi-Level Adaptive Token Refinement for Few-Shot Transformers (BATR-FST), a two-stage approach that progressively improves token representations and maintains a robust inductive bias for few-shot classification. During the pre-training phase, Masked Image Modeling (MIM) provides Vision Transformers (ViTs) with transferable patch-level representations by recreating masked image regions, providing a robust basis for subsequent adaptation. In the meta-fine-tuning phase, BATR-FST incorporates a Bi-Level Adaptive Token Refinement module that utilizes Token Clustering to capture localized interactions, Uncertainty-Aware Token Weighting to prioritize dependable features, and a Bi-Level Attention mechanism to balance intra-cluster and inter-cluster relationships, thereby facilitating thorough token refinement. Furthermore, Graph Token Propagation ensures semantic consistency between support and query instances, while a Class Separation Penalty preserves different class borders, enhancing discriminative capability. Extensive experiments on three benchmark few-shot datasets demonstrate that BATR-FST achieves superior results in both 1-shot and 5-shot scenarios and improves the few-shot classification via transformers.
\end{abstract}


\section{Introduction}
Recent advancements in deep learning have allowed the development of more complex models trained on large and diverse datasets. With access to abundant training data, deep learning has achieved impressive results in computer vision tasks such as image classification, object detection, and segmentation \cite{LeCun2015, Krizhevsky2012, He2017}. However, many real-world applications struggle to use deep learning because large labeled datasets are scarce or expensive. Industries such as healthcare and quality control often work with small datasets or require extensive resources for labeling, making it challenging to gather enough data for traditional deep-learning models. Consequently, effectively using small datasets has become an important research area, leading to developing few-shot learning as a promising solution. Inspired by the human ability to recognize new objects from just a few examples, few-shot learning aims to develop models that generalize well with minimal labeled data. Despite its potential, few-shot learning often struggles with overfitting and poor generalization, especially when training data is scarce.
At the same time, Vision Transformers \cite{Dosovitskiy2021, wang2021not} have proven to be powerful models in computer vision, often outperforming traditional convolutional neural networks by effectively capturing local and global image features. Combining Vision Transformers with few-shot learning presents a valuable opportunity to leverage their strong feature extraction capabilities. However, a key challenge lies in effectively adapting Vision Transformers to few-shot scenarios, mainly because they lack the built-in inductive biases that convolutional neural networks possess.

Several studies have attempted to address the challenges of adapting Vision Transformers (ViTs) to Few-Shot Learning (FSL). For example, SUN \cite{dong2022self} focuses on improving intertoken dependency learning in Vision Transformers by applying intensive supervision at specific locations to address the lack of inductive bias in ViTs. Similarly, FewTURE \cite{hiller2022rethinking} divides input samples into segments, encoding them with ViTs to establish semantic relationships and employing masked image modeling (MIM) to reduce the adverse effects of noisy annotations. Furthermore, TATM \cite{li2024tatm} proposes a framework to extract knowledge from pretrained transformers and adapt it to downstream FSL tasks. Although these methods represent necessary steps forward, they remain limited in their ability to fully exploit the potential of Vision Transformers in few-shot scenarios. Persistent challenges, including overfitting, poor generalization, and diverse task distributions, underscore the need for more effective solutions tailored to FSL constraints.

In this paper, we introduce Bi-Level Adaptive Token Refinement for Few-Shot Transformers (BATR-FST), a framework designed to tackle few-shot classification challenges using Vision Transformers (ViTs) with a progressive token refinement process. BATR-FST consists of two stages: pretraining and meta-finetuning. In pretraining, Masked Image Modeling (MIM) \cite{zhouimage} trains the ViT, enabling it to capture local features and global representations. During meta-finetuning, the Bi-Level Graph Token Attention Transformer Block refines token representations through token selection, uncertainty estimation, intra-cluster attention, inter-cluster attention, and graph token propagation. This hierarchical refinement ensures robust feature aggregation, effectively handling few-shot scenarios. Unlike current transformer-based few-shot learning methods, which often rely on static token matching or simplistic similarity mechanisms, our approach dynamically refines token representations to effectively balance global context with local feature interactions. Extensive experiments on mini-ImageNet, tiered-ImageNet, and CIFAR-FS demonstrate the effectiveness of our framework, establishing a strong baseline for future advancements in transformer-based few-shot learning. Our main contributions are outlined as follows:
\begin{itemize}
    \item We propose a novel Bi-Level Adaptive Token Refinement mechanism for few-shot transformers that enhances token representations at local and global levels, significantly improving feature extraction and generalization in data-scarce scenarios.
    \item We develop an Uncertainty-Aware Token Weighting strategy and a Graph Token Propagation module to evaluate token reliability and model dependencies, fostering effective semantic interactions between support and query tokens.
    \item A Class Separation Penalty, inspired by contrastive learning, is designed to enforce intra-class compactness and inter-class separability, enhancing the discriminative power of the model.
    \item Extensive experiments on mini-ImageNet, tiered-ImageNet, and CIFAR-FS validate our framework, providing a strong baseline for 1-shot and 5-shot tasks in few-shot learning.
\end{itemize}

\section{Related Work} \label{Related_Work}

\subsection{Few-Shot Learning}
Few-shot learning (FSL) aims to train models that generalize effectively from limited labeled examples, addressing challenges in data-scarce scenarios. Meta-learning, or learning-to-learn, underpins FSL and encompasses metric-based \cite{snell2017prototypical,vinyals2016matching, ye2020few} and optimization-based approaches \cite{finn2017model, antoniou2018train}. 
Metric-based methods, such as ProtoNets \cite{snell2017prototypical}, learn embedding spaces where classification relies on distances to class prototypes. Enhancements like FEAT \cite{ye2020few} utilize set-to-set functions to refine embeddings for better adaptability across datasets.
Optimization-based methods, such as MAML\cite{finn2017model} and Reptile\cite{zhang2021rethinking}, enable models to adapt to new tasks by learning optimal initial parameters.
Recent research has explored transformer-based architectures for FSL due to their strong representational capabilities \cite{Dosovitskiy2021, wang2023few, du2024metakernel}. However, most existing transformer-based methods rely on fixed CNN-based feature extractors, limiting their adaptability and underutilizing the full potential of Vision Transformers (ViTs). To address these limitations, our approach integrates tokenization, attention mechanisms, and graph-based representations to dynamically exploit both local and global features, enhancing performance and adaptability in few-shot scenarios.

\subsection{Vision Transformers in Few-Shot Learning}

Vision Transformers (ViTs) \cite{Dosovitskiy2021}, adapted from NLP architectures \cite{vaswani2017attention}, have become a robust framework for computer vision. Enhanced variants like CeiT \cite{yuan2021incorporating} and NesT \cite{zhang2022nested} improve performance with techniques such as knowledge distillation. However, ViTs still depend on large datasets, limiting their use in few-shot learning (FSL). To address this, self-supervised methods BEiT \cite{bao2021beit} leverage masking for improved generalization, and MFGN \cite{yu2022masked} demonstrates the utility of feature augmentation in FSL. Many meta-learning algorithms for FSL still rely on CNN-based feature extractors \cite{wang2023few, nguyen2023pac, fu2024prototype}, which limit flexibility and underutilize the full potential of ViTs. Moreover, ViTs lack inductive biases such as spatial locality and translation equivariance, complicating their application to FSL. Studies \cite{dong2022self, hiller2022rethinking, li2024tatm} have introduced strategies to address these issues, including self-supervised learning from few-shot datasets. Masked self-supervised methods \cite{zhou2021ibot} enable ViTs to capture both local features and global context. Drawing inspiration from GraphViT\cite{he2023generalization}, which extends Vision Transformers with subgraph tokens for graph-based learning, we integrate tokenization, attention mechanisms, and graph representations to enhance ViTs for few-shot learning and domain adaptation in data-scarce settings.

\section{Methodology}

This section explains the proposed Bi-Level Adaptive Token Refinement framework for Few-Shot Transformers (BATR-FST). Section~\ref{sec:problem_definition} defines the few-shot learning problem and establishes its formal setup. Section~\ref{sec:Overview} presents an overview of the framework, outlining its key components and processes. Sections~\ref{MIM} and~\ref{Bi-Level_Adaptive_Token} describe the pre-training stage and the Bi-Level Adaptive Token modules essential for effective token refinement. Lastly, Section~\ref{Class_Separation} introduces the class separation penalty and summarizes the formulation of the overall loss function. 

\begin{figure*}[t!]
    \centering
    \includegraphics[width=1.0\textwidth]{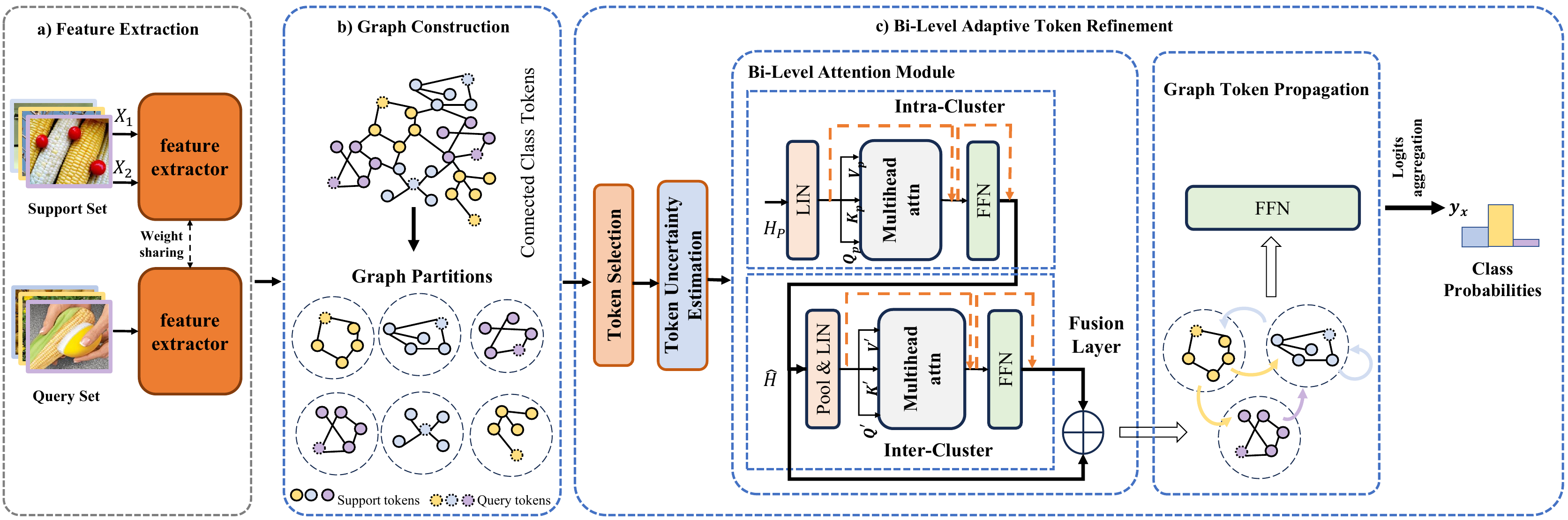}
    \caption{The framework of our Bi-Level Adaptive Token Refinement method utilizes a ViT-Small architecture as the feature extractor.}
    \label{fig:proposed}
\end{figure*}

\subsection{Problem Definition} \label{sec:problem_definition}
\label{subsec:problem_definition}
Few-shot classification aims to train a model capable of accurately predicting labels for new samples, even when provided with only a few labeled examples per class. In this setting, tasks are designed to simulate the sparse data characteristics often encountered in real-world scenarios. Each task \( \mathcal{T} \) is composed of a support set \( \mathcal{S} \) and a query set \( \mathcal{Q} \). The support set \( \mathcal{S} \) contains \( N \) classes with \( K \) labeled examples per class, i.e., \( \mathcal{S} = \{(x_i, y_i)\}_{i=1}^{N \times K} \), while the query set \( \mathcal{Q} \) consists of \( Q \) unlabeled examples, i.e., \( \mathcal{Q} = \{x_i\}_{i=N \times K + 1}^{N \times K + Q} \). Crucially, the classes in the support and query sets are disjoint, such that \( \mathcal{C}_S \cap \mathcal{C}_Q = \emptyset \), where \( \mathcal{C}_S \) and \( \mathcal{C}_Q \) denote the class sets of \( \mathcal{S} \) and \( \mathcal{Q} \), respectively. In the \( N \)-way \( K \)-shot setting, \( N \) classes are randomly selected for each task, with \( K \) labeled samples assigned to the support set and \( Q \) samples allocated to the query set. The objective is to predict the labels \( \hat{y}_i \) for query samples \( x_i \in \mathcal{Q} \) using the limited labeled data in \( \mathcal{S} \). 

\subsection{Overview of Framework} \label{sec:Overview}
We propose a two-stage framework for few-shot learning, the Bi-Level Adaptive Token Refinement Framework for Few-Shot Transformers (BATR-FST). In the pre-training stage, Masked Image Modeling (MIM) trains a Vision Transformer (ViT) by reconstructing randomly masked image patches, enabling the model to learn robust patch-level features and generate semantically rich token embeddings. In the meta-finetuning stage, the pre-trained ViT is refined using the Bi-Level Adaptive Token Refinement Module (BATR), which integrates Token Clustering, Uncertainty-Aware Token Weighting, and the Bi-Level Attention Mechanism to adapt the model for few-shot tasks. Graph Token Propagation and Class Separation Penalty are incorporated to enhance contextual representation and promote discriminative learning. This comprehensive approach ensures that BATR-FST effectively integrates semantic and local features, focusing on the most reliable tokens to maximize classification accuracy in few-shot scenarios. The overall process of the proposed framework is illustrated in Fig.~\ref{fig:proposed}.

\subsection{Stage I: MIM-Based Vision Transformer Pre-Training} \label{MIM}

In the initial stage, we pre-train a Vision Transformer (ViT) using a Masked Image Modeling (MIM) strategy inspired by Masked Autoencoders. Given an input image \( \mathbf{x} \in \mathbb{R}^{H \times W \times C} \), the image is partitioned into \( L \) non-overlapping patches. A class token \( \mathbf{z}_{\mathrm{cls}} \) is prepended to the sequence of patch tokens, and learnable positional embeddings are added to incorporate spatial information. We randomly mask 75\% of these patches, resulting in a partially observed version \( \mathbf{x}_{\mathrm{masked}} \).

The ViT encoder \( f_{\mathrm{ViT}}(\mathbf{x}; \theta_{\mathrm{emb}}) \) is trained to reconstruct the masked patches by minimizing the Mean Squared Error (MSE) loss:
\begin{equation}
\mathcal{L}_{\mathrm{MIM}} = \frac{1}{|\mathcal{M}|} \sum_{p \in \mathcal{M}} \left\| \mathbf{x}_p - \hat{\mathbf{x}}_p \right\|^2,
\end{equation}
where \( \mathcal{M} \) denotes the set of masked patches and \( \hat{\mathbf{x}}_p \) is the reconstructed patch at position \( p \). This objective compels the Transformer to learn robust patch-level features that capture diverse visual patterns, even under substantial occlusion.
Upon convergence of the MIM pre-training, each unmasked image \( \mathbf{x}_i \) is fed into the ViT encoder to extract token embeddings:
\begin{equation}
\mathbf{z}_i^0 \in \mathbb{R}^{(L+1) \times D},
\end{equation}
where \( D \) is the dimensionality of each token embedding. The embedding \( \mathbf{z}_i^0 \) comprises one class token \( \mathbf{z}_{i,\mathrm{cls}}^0 \) and \( L \) patch tokens \( \{ \mathbf{z}_{i,1}^0, \dots, \mathbf{z}_{i,L}^0 \} \). These token embeddings form the foundational representations for the subsequent few-shot meta-learning stage.

\subsection{Stage II: Bi-Level Adaptive Token Refinement for Few-Shot Learning} \label{Bi-Level_Adaptive_Token}

The second stage employs the Bi-Level Adaptive Token Refinement Module (BATR), which integrates complementary mechanisms to refine the pre-trained representations for robust generalization.
For each support image, two augmented views \( \mathbf{x}_1 \) and \( \mathbf{x}_2 \) are generated using random cropping and color jittering. Passing these through the ViT encoder produces token embeddings \( \mathbf{z}_1 \) and \( \mathbf{z}_2 \), consisting of patch tokens \( \mathbf{z}_{1,j}, \mathbf{z}_{2,j} \) (\( j = 1, \dots, L \)) and class tokens \( \mathbf{z}_{1,\mathrm{cls}}, \mathbf{z}_{2,\mathrm{cls}} \). 

\subsubsection{Graph Construction and Token Clustering}

Support \( X_S \) and query \( X_Q \) token embeddings are combined as nodes in a fully connected graph, with edge similarities defined as \( e_{ij} = \text{sim}(z_i, z_j) \). The adjacency matrix \( A \) is normalized to obtain attention scores:
\begin{equation}
A_{ij} = \frac{\exp(e_{ij})}{\sum_k \exp(e_{ik})}.
\end{equation}
To manage computational complexity and enhance localized coherence, the graph is partitioned into \( K \) clusters \( G_1, \dots, G_K \) using the Metis algorithm~\cite{karypis1998fast}.

\subsubsection{Uncertainty-Aware Token Weighting}

Few-shot tasks often produce noisy or ambiguous tokens. To quantify token reliability, we estimate each token's uncertainty using Monte Carlo Dropout. Specifically, we run the ViT with stochastic dropout for \( T \) forward passes, obtaining token embeddings \( \{ \mathbf{z}_p^{(t)} \}_{t=1}^T \). The variance of each token embedding \( \mathbf{z}_p \) is computed as:
\begin{equation}
\scalebox{0.9}{$
\mathrm{Var}(\mathbf{z}_p) = \frac{1}{T} \sum_{t=1}^T \left\| \mathbf{z}_p^{(t)} - \overline{\mathbf{z}}_p \right\|^2,
\quad
\overline{\mathbf{z}}_p = \frac{1}{T} \sum_{t=1}^T \mathbf{z}_p^{(t)}.
$}
\end{equation}
The maximum variance across all tokens normalizes this variance to obtain a final uncertainty score \( \tilde{u}_p \in [0, 1] \):
\begin{equation}
\tilde{u}_p = \frac{\mathrm{Var}(\mathbf{z}_p)}{\max_{q} \mathrm{Var}(\mathbf{z}_q)}.
\end{equation}
A larger \( \tilde{u}_p \) indicates lower reliability for token \( p \).

We evaluate token importance using uniqueness (\( \gamma_p \), attention received by token \( p \)) and broadcast influence (\( \psi_p \), attention distributed by \( p \)), combined as:
\begin{equation}
\eta_p = \gamma_p \times \psi_p.
\end{equation}
Tokens with high \( \eta_p \) are influential, while those with low \( \eta_p \) or high uncertainty \( \tilde{u}_p \) are down-weighted or pruned. When constructing the importance mask \( \mathbf{M} \), synergy scores are further scaled by \( (1 - \tilde{u}_p) \) or \( \eta_p \), prioritizing reliable and impactful tokens while de-emphasizing uncertain ones.

\subsubsection{Bi-Level Attention Mechanism}

The Bi-Level Attention Mechanism captures both local and global token interactions through two stages: \textit{Intra-Cluster Attention} and \textit{Inter-Cluster Attention}.

\textit{Intra-Cluster Attention} processes tokens within each cluster using a Transformer block to model localized interactions. Token embeddings are updated via self-attention:

\begin{equation}
\scalebox{0.8}{$\mathbf{H}_p^1 = \mathrm{FFN}\left( \mathrm{Softmax}\left( \frac{\mathbf{Q}_p \mathbf{K}_p^\top}{\sqrt{d}} \right) \mathbf{V}_p \right)$}
\end{equation}
where queries (\( \mathbf{Q}_p \)), keys (\( \mathbf{K}_p \)), and values (\( \mathbf{V}_p \)) are derived from \( \mathbf{H}_p^0 \), and \( \mathrm{FFN} \) is a feed-forward network. The output \( \mathbf{H}_p^1 \) is pooled into \( \mathbf{c}_p \), summarizing the cluster's local context.

\textit{Inter-Cluster Attention} processes the pooled cluster embeddings \( \mathbf{C} = \{\mathbf{c}_1, \dots, \mathbf{c}_K\} \) to capture global relationships:
\begin{equation}
\scalebox{0.8}{$\mathbf{C}' = \mathrm{FFN}\left( \mathrm{Softmax}\left( \frac{\mathbf{Q}' (\mathbf{K}')^\top}{\sqrt{d}} \right) \mathbf{V}' \right)$}
\end{equation}
where \( \mathbf{Q}', \mathbf{K}', \mathbf{V}' \) are derived from \( \mathbf{C} \). The refined global embedding \( \mathbf{C}' \) is fused with \( \mathbf{H}_p^1 \) to enrich token representations:
\begin{equation}
\scalebox{0.8}{$\mathbf{H}_p^2 = \left[ \mathbf{H}_p^1 \;||\; \mathbf{c}_p' \right] \mathbf{W}_{\mathrm{fuse}}$}
\end{equation}
where \( \mathbf{W}_{\mathrm{fuse}} \) is a learnable projection matrix. This integration balances fine-grained local refinement and global contextual information, enabling enriched feature representation for downstream tasks.

\subsubsection{Graph Token Propagation}

Following the refinement from the Bi-Level Attention Mechanism, the Graph Token Propagation module enables tokens to interact globally across the entire graph, further enhancing their contextual representation. The refined embeddings from the Bi-Level Attention, denoted as \( \mathbf{H} \in \mathbb{R}^{N \times d} \), where \( N \) is the total number of tokens and \( d \) is the embedding dimension, serve as input to this module. Tokens are updated through a global attention mechanism:
\begin{equation}
\scalebox{0.8}{$\mathbf{H}' = \text{Softmax}\left( \frac{\mathbf{H} \mathbf{W}_Q (\mathbf{H} \mathbf{W}_K)^\top}{\sqrt{d}} \right) \mathbf{H} \mathbf{W}_V$}
\end{equation}

where \( \mathbf{W}_Q, \mathbf{W}_K, \mathbf{W}_V \in \mathbb{R}^{d \times d} \) are learnable projection matrices. The output embeddings, \( \mathbf{H}' \), are further refined through a feed-forward network (FFN) to produce globally coherent representations. These propagated embeddings compute class probabilities, ensuring effective information flow and consistency across support and query tokens.

\subsection{Class Separation Penalty} \label{Class_Separation}

To promote intra-class consistency and inter-class distinctness in few-shot settings, we incorporate a Class Separation Penalty \( \mathcal{L}_{\mathrm{sep}} \). For any pair of tokens (or class embeddings) \( \mathbf{z}_i \) and \( \mathbf{z}_j \):
\begin{equation}
\mathcal{L}_{\mathrm{sep}} = \frac{
  \sum_{(i,j) \in \mathcal{P}_+} \mathrm{dist}(\mathbf{z}_i, \mathbf{z}_j)
}{
  \sum_{(i,j) \in \mathcal{P}_-} \mathrm{dist}(\mathbf{z}_i, \mathbf{z}_j)
},
\end{equation}
where \( \mathcal{P}_+ \) denotes pairs of tokens from the same class and \( \mathcal{P}_- \) denotes pairs of tokens from different classes. The distance metric \( \mathrm{dist}(\mathbf{z}_i, \mathbf{z}_j) \) typically represents the Euclidean distance between tokens \( \mathbf{z}_i \) and \( \mathbf{z}_j \). This ratio encourages the model to minimize distances within the same class while maximizing distances between different classes, thereby enhancing the discriminative power of the model in data-scarce regimes.

\textit{Meta-Learning Loss and Final Prediction} In the meta-training phase, let \( \hat{y}_i \) be the predicted distribution for token (or image) \( i \), and \( y_i \) its ground-truth label. The overall meta-training objective combines the standard cross-entropy loss \( \mathcal{L}_{\mathrm{CE}} \) with the separation penalty \( \mathcal{L}_{\mathrm{sep}} \):
\begin{equation}
\mathcal{L}_{\mathrm{meta}} = \alpha\,\mathcal{L}_{\mathrm{CE}} + \beta\,\mathcal{L}_{\mathrm{sep}},
\end{equation}
where \( \alpha \) and \( \beta \) are hyperparameters determined via validation. This objective ensures the model correctly classifies the samples and maintains distinct class boundaries, enhancing performance in few-shot learning scenarios.

\section{Experiments}

\subsection{Datasets}

\textbf{The mini-ImageNet dataset}, proposed by \cite{vinyals2016matching}, has 100 categories, each comprising 600 images, resulting in a total of 60,000 samples. Following previous studies, we partitioned the dataset into 64 categories for training, 16 for validation, and 20 for testing.

\textbf{The tiered-ImageNet dataset}, introduced by \cite{ren2018meta}, has 34 broad categories, each containing 608 specific classes. In our studies, we adhered to the conventional method by employing 20 categories for training, 6 for validation, and 8 for testing.

\textbf{The CIFAR-FS dataset}, introduced by Bertinetto et al.~\cite{bertinetto2019metalearning}, is a few-shot learning benchmark derived from CIFAR-100. It contains 100 classes organized into 20 superclasses, split into 60 classes (12 superclasses) for training, 20 classes (4 superclasses) for validation, and 20 classes (4 superclasses) for testing, enabling diverse generalization evaluations.

\subsection{Implementation Details}

Our framework uses the Vision Transformer (ViT-S) for pre-training and meta-finetuning. In the pre-training phase, we apply Masked Image Modeling (MIM) on mini-ImageNet, tiered-ImageNet, and CIFAR-FS datasets for 1,700 epochs, resizing images to \(224 \times 224\) pixels and masking 75\% of patches. We use the AdamW optimizer with a batch size of 128, learning rate \(3 \times 10^{-4}\), and cosine annealing decay, along with data augmentation (horizontal flipping, color jittering). During meta-finetuning, the pre-trained ViT is refined for few-shot learning using SGD with learning rate \(1 \times 10^{-4}\) and weight decay \(4 \times 10^{-4}\) for 80 epochs per dataset, with 100 training and 50 validation tasks. We evaluate on 1,000 tasks, using 5-shot (5 support and 15 query samples) and 1-shot (1 support and 15 query samples) settings. Inner-loop optimization is performed with SGD (learning rate 0.1) for 35 iterations, and a similarity matrix masking mechanism mitigates overfitting. Token refinement uses \(k_{\text{local}} = 20\) for local clusters and \(k_{\text{global}} = 20\) for global token selection. Bi-Level Attention is configured with 20 clusters and 50\% attention sparsification, with adaptive temperature parameters initialized at 0.1. The class separation penalty weight \(\beta\) is 0.5, and the meta-learning loss balance weight is 0.4. During inference, query tokens are processed through the Bi-Level Attention Mechanism and Graph Token Propagation, improving performance on 1-shot and 5-shot tasks.

\subsection{Results on Benchmark Datasets}
We evaluated BATR-FST on three standard few-shot learning benchmarks: \textit{Mini-ImageNet}, \textit{Tiered-ImageNet}, and \textit{CIFAR-FS}, utilizing both 5-way 1-shot and 5-shot settings. The performance of BATR-FST was compared against recent methods categorized by their backbone architectures: \textit{ResNet-12}, \textit{WRN-28-10}, and \textit{Vision Transformers (ViT)}, as detailed in Tables \ref{tab:few_shot_comparison} and \ref{tab:cifar_fs_comparison}.
On the \textit{Mini-ImageNet} dataset, BATR-FST achieves superior performance, attaining 70.36\% ± 0.08 accuracy in the 1-shot and 86.50\% ± 0.30 in the 5-shot classification settings, surpassing both \textit{ResNet-12} and \textit{WRN-28-10}-based methods. These gains are due to the Bi-Level Attention Mechanism and Graph Token Propagation module, which model fine-grained and holistic interactions, respectively, and the Uncertainty-Aware Token Weighting strategy that enhances feature aggregation by prioritizing reliable tokens. Compared to recent ViT-based methods such as TATM \cite{li2024tatm}, and QSFormer \cite{wang2023few}, BATR-FST demonstrates better generalization and efficient parameter usage (22M parameters).
On the \textit{Tiered-ImageNet} dataset, BATR-FST achieves 73.80\% ± 0.50 accuracy in the 1-shot setting and 88.20\% ± 0.25 in the 5-shot setting. Leveraging the ViT-S backbone with 22M parameters, BATR-FST maintains competitive accuracy compared to \textit{WRN-28-10}-based methods with 36.5M parameters, highlighting its efficiency and scalability.
On the \textit{CIFAR-FS} dataset, BATR-FST achieves 79.81\% ± 0.75 in the 1-shot and 90.70\% ± 0.25 in the 5-shot classification settings, outperforming \textit{ResNet-12} and \textit{ViT-S}-based methods. BATR-FST surpasses QSFormer \cite{wang2023few} in 1-shot accuracy and achieves comparable performance in 5-shot, demonstrating its effectiveness in more challenging low-data settings.
The comparison experiments collectively indicate that our method enhances classification accuracy in both 1-shot and 5-shot tasks, delivering highly competitive results.

\begin{table*}[ht]
\scriptsize

\centering
\caption{Comparison of few-shot learning methods on mini-ImageNet and tiered-ImageNet (5-way classification). Accuracy (\%) with 95\% confidence intervals is reported. The best results are in bold. }

\label{tab:few_shot_comparison}

\begin{tabular}{@{}llccccc@{}}

\toprule
\textbf{Model} & \textbf{Backbone} & \textbf{\#Params} & \multicolumn{2}{c}{\textbf{Mini-ImageNet}} & \multicolumn{2}{c}{\textbf{Tiered-ImageNet}} \\
\cmidrule(lr){4-5} \cmidrule(lr){6-7}
 &  &  & \textbf{1-shot} & \textbf{5-shot} & \textbf{1-shot} & \textbf{5-shot} \\ 
\midrule
ProtoNets \cite{snell2017prototypical} & ResNet-12 & 12.4M & 60.37 $\pm$ 0.83 & 78.02 $\pm$ 0.57 & 65.65 $\pm$ 0.92 & 83.40 $\pm$ 0.65 \\
Meta-Baseline \cite{chen2021meta} & ResNet-12 & 12.4M & 63.17 $\pm$ 0.23 & 79.26 $\pm$ 0.17 & 68.62 $\pm$ 0.27 & 83.29 $\pm$ 0.18 \\
MCL \cite{liu2022learning} & ResNet-12 & 12.4M & 67.51 $\pm$ 0.20 & 83.99 $\pm$ 0.20 & 72.01 $\pm$ 0.20 & 86.02 $\pm$ 0.20 \\
Meta DeepBDC \cite{xie2022joint} & ResNet-12 & 12.4M & 67.34 $\pm$ 0.43 & 84.46 $\pm$ 0.28 & 72.34 $\pm$ 0.49 & 87.31 $\pm$ 0.32 \\
Meta-HP \cite{zhang2023meta} & ResNet-12 & 12.4M & 62.49 $\pm$ 0.80 & 77.12 $\pm$ 0.62 & 68.26 $\pm$ 0.72 & 82.91 $\pm$ 0.36 \\
SoSN \cite{zhang2022multi} & ResNet-12 & 12.4M & 58.26 $\pm$ 0.87 & 73.20 $\pm$ 0.68 & 58.62 $\pm$ 0.92 & 75.19 $\pm$ 0.79 \\
CME \cite{zhou2023learning} & ResNet-12 & 12.4M & 63.01 $\pm$ 0.80 & 79.78 $\pm$ 0.14 & 67.18 $\pm$ 0.23 & 82.84 $\pm$ 0.31 \\
CAML \cite{gao2022curvature} & ResNet-12 & 12.4M & 63.13 $\pm$ 0.41 & 81.04 $\pm$ 0.39 & 68.46 $\pm$ 0.56 & 83.84 $\pm$ 0.40 \\
CADS \cite{zhang2024few} & ResNet-12 & 12.4M & 66.56 $\pm$ 0.19 & 82.74 $\pm$ 0.13 & 72.04 $\pm$ 0.22 & 86.47 $\pm$ 0.15 \\ 
PBML \cite{fu2024prototype} & ResNet-12 & 12.4M & 63.60 $\pm$ 0.70 & 81.94 $\pm$ 0.44 & 70.64 $\pm$ 0.72 & 85.39 $\pm$ 0.40 \\ 
\midrule
FEAT \cite{ye2020few} & WRN-28-10 & 36.5M & 65.10 $\pm$ 0.20 & 81.11 $\pm$ 0.14 & 70.41 $\pm$ 0.23 & 84.38 $\pm$ 0.16 \\
SImPa \cite{nguyen2023pac} & WRN-28-10 & 36.5M & 62.85 $\pm$ 0.56 & 77.65 $\pm$ 0.50 & 70.26 $\pm$ 0.35 & 80.15 $\pm$ 0.28 \\
PBML \cite{fu2024prototype} & WRN-28-10 & 36.5M & 65.85 $\pm$ 0.73 & 83.04 $\pm$ 0.43 & 73.29 $\pm$ 0.76 & 86.75 $\pm$ 0.40 \\
\midrule
SUN \cite{dong2022self} & NesT & 12.8M & 66.54 $\pm$ 0.45 & 82.09 $\pm$ 0.30 & 72.93 $\pm$ 0.50 & 86.70 $\pm$ 0.33 \\
SUN \cite{dong2022self} & Visformer & 12.5M & 67.80 $\pm$ 0.45 & 83.25 $\pm$ 0.30 & 72.99 $\pm$ 0.50 & 86.74 $\pm$ 0.33 \\
FewTURE \cite{hiller2022rethinking} & ViT-S & 22M & 68.02 $\pm$ 0.88 & 84.51 $\pm$ 0.53 & 72.96 $\pm$ 0.92 & 86.43 $\pm$ 0.67 \\
SImPa \cite{nguyen2023pac} & NesT ViT & 12.8M & 68.15 $\pm$ 0.82 & 82.96 $\pm$ 0.55 & 73.38 $\pm$ 0.93 & 86.87 $\pm$ 0.54 \\
QSFormer \cite{wang2023few} & NesT ViT & 12.8M & 65.97 $\pm$ 0.91 & 80.58 $\pm$ 0.50 & 73.28 $\pm$ 0.64 & 87.19 $\pm$ 0.73 \\
TATM \cite{li2024tatm} & ViT-S & 22M & 68.89 $\pm$ 0.82 & 85.89 $\pm$ 0.25 & 73.02 $\pm$ 0.64 & 87.74 $\pm$ 0.33 \\ 
\textbf{BATR-FST (Ours)} & ViT-S & 22M & \textbf{70.36 $\pm$ 0.08} & \textbf{86.50 $\pm$ 0.30} & \textbf{73.80 $\pm$ 0.50} & \textbf{88.20 $\pm$ 0.25} \\
\bottomrule
\end{tabular}%

\end{table*}

\begin{table}[ht]
\centering
\caption{Comparison of few-shot learning methods on CIFAR-FS (5-way classification). Accuracy (\%) with 95\% confidence intervals is reported. Best results in each column are in bold.}

\label{tab:cifar_fs_comparison}
\resizebox{\columnwidth}{!}{%
\begin{tabular}{@{}llccc@{}}
\toprule
\textbf{Model} & \textbf{Backbone} & \textbf{\#Params} & \multicolumn{2}{c}{\textbf{CIFAR-FS}} \\
\cmidrule(lr){4-5}
 &  &  & \textbf{1-shot} & \textbf{5-shot} \\ 
\midrule
ProtoNets \cite{snell2017prototypical} & ResNet-12 & 12.4M & 72.20 $\pm$ 0.70 & 83.50 $\pm$ 0.50 \\
Meta-NVG \cite{zhang2021meta} & ResNet-12 & 12.4M & 74.63 $\pm$ 0.91 & 86.45 $\pm$ 0.59 \\
   
CME \cite{zhou2023learning} & ResNet-12 & 12.4M & 72.63 $\pm$ 0.31 & 85.88 $\pm$ 0.15 \\
GCLR-SVM \cite{zhong2022graph} & ResNet-12 & 12.4M & 74.10 $\pm$ 0.70 & 87.10 $\pm$ 0.50 \\
CADS \cite{zhang2024few} & ResNet-12 & 12.4M & 73.23 $\pm$ 0.21 & 87.67 $\pm$ 0.14 \\ 
\midrule
FewTURE \cite{hiller2022rethinking} & ViT-S & 22M & 76.10 $\pm$ 0.88 & 86.14 $\pm$ 0.64 \\
TATM \cite{li2024tatm} & ViT-S & 22M & 76.50 $\pm$ 0.86 & 87.96 $\pm$ 0.32 \\ 
MetaKernel \cite{du2024metakernel} & NesT ViT & 12.8M & 79.67 $\pm$ 0.62 & 89.63 $\pm$ 0.58 \\
SImPa \cite{nguyen2023pac} & NesT ViT & 12.8M & 78.33 $\pm$ 0.76 & 90.31 $\pm$ 0.50 \\
QSFormer \cite{wang2023few} & NesT ViT & 12.8M & 79.40 $\pm$ 0.59 & \textbf{90.73 $\pm$ 0.73} \\

\textbf{BATR-FST (Ours)} & ViT-S & 22M & \textbf{79.81 $\pm$ 0.75} & \textbf{90.70 $\pm$ 0.25} \\
\bottomrule
\end{tabular}%
}
\end{table}

\subsection{Visualization Analysis}
\begin{figure}[h!]
    \centering
    \includegraphics[width=\columnwidth]{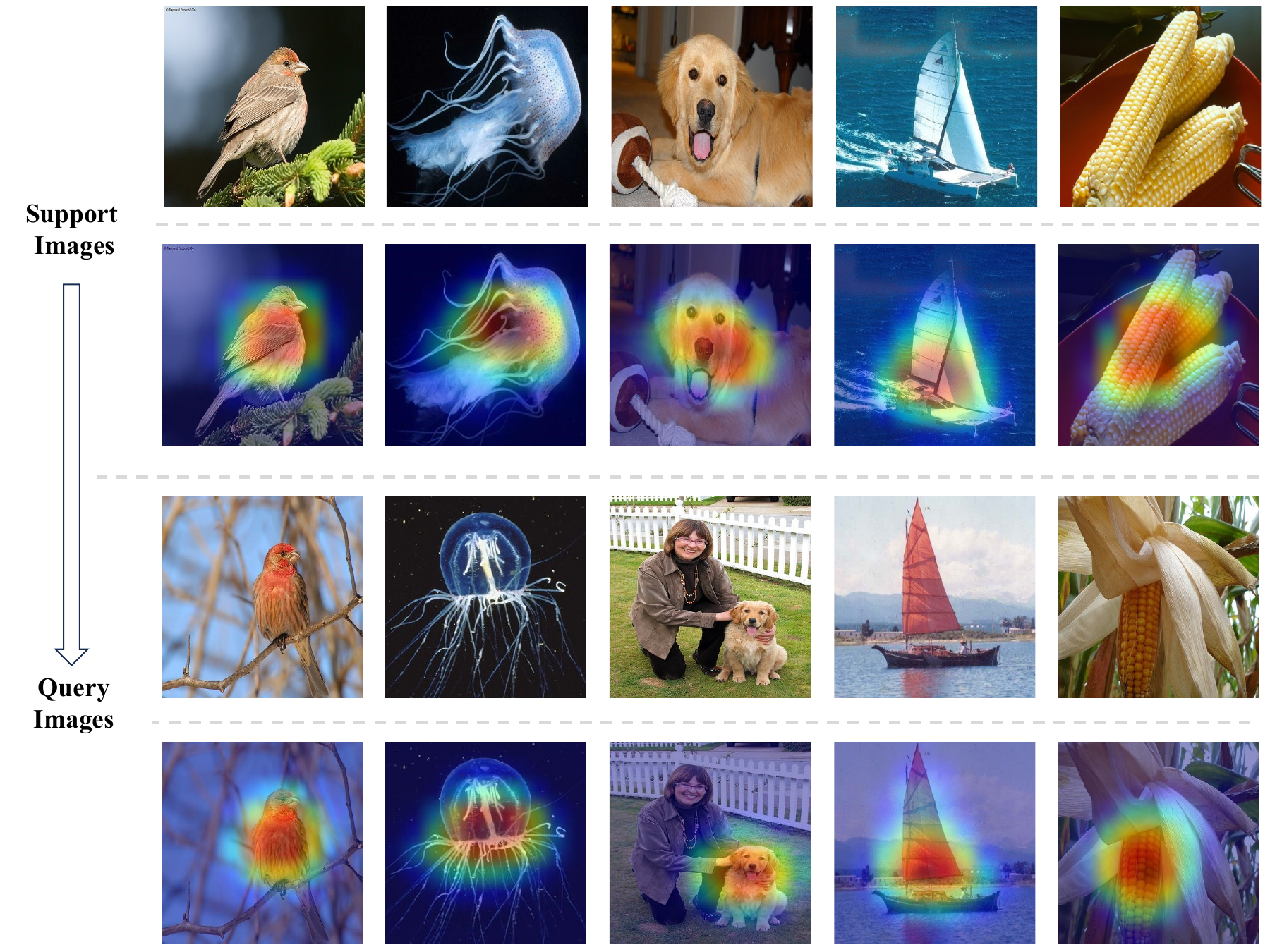}
    \caption{The Grad-Cam visualization of our method on the mini-Imagenet. Each column in both groups belongs to the same class.}
    \label{fig:gradcamvis}
\end{figure}

Figure~\ref{fig:gradcamvis} presents the results of our Grad-CAM visualization analysis on a 5-shot task from the mini-ImageNet dataset. We conducted 35 inner-loop iterations on the support set associated with the task and used the fine-tuned weights to generate predictions on the query set for Grad-CAM visualization. The figure demonstrates the effectiveness of our method in identifying and focusing on the most relevant regions of the input images that contribute to correct classification. Specifically, the third column of Figure~\ref{fig:gradcamvis} highlights the ability of our approach to reliably recognize the "dog" class within a sample containing multiple categories. This underscores the model's robustness in accurately identifying key features for classification while effectively ignoring irrelevant background elements that do not contribute to the classification objective.

\subsection{Parameter Analysis}

\begin{figure}[ht]
    \centering
    \begin{subfigure}[b]{0.50\columnwidth}
        \centering
        \includegraphics[width=\linewidth]{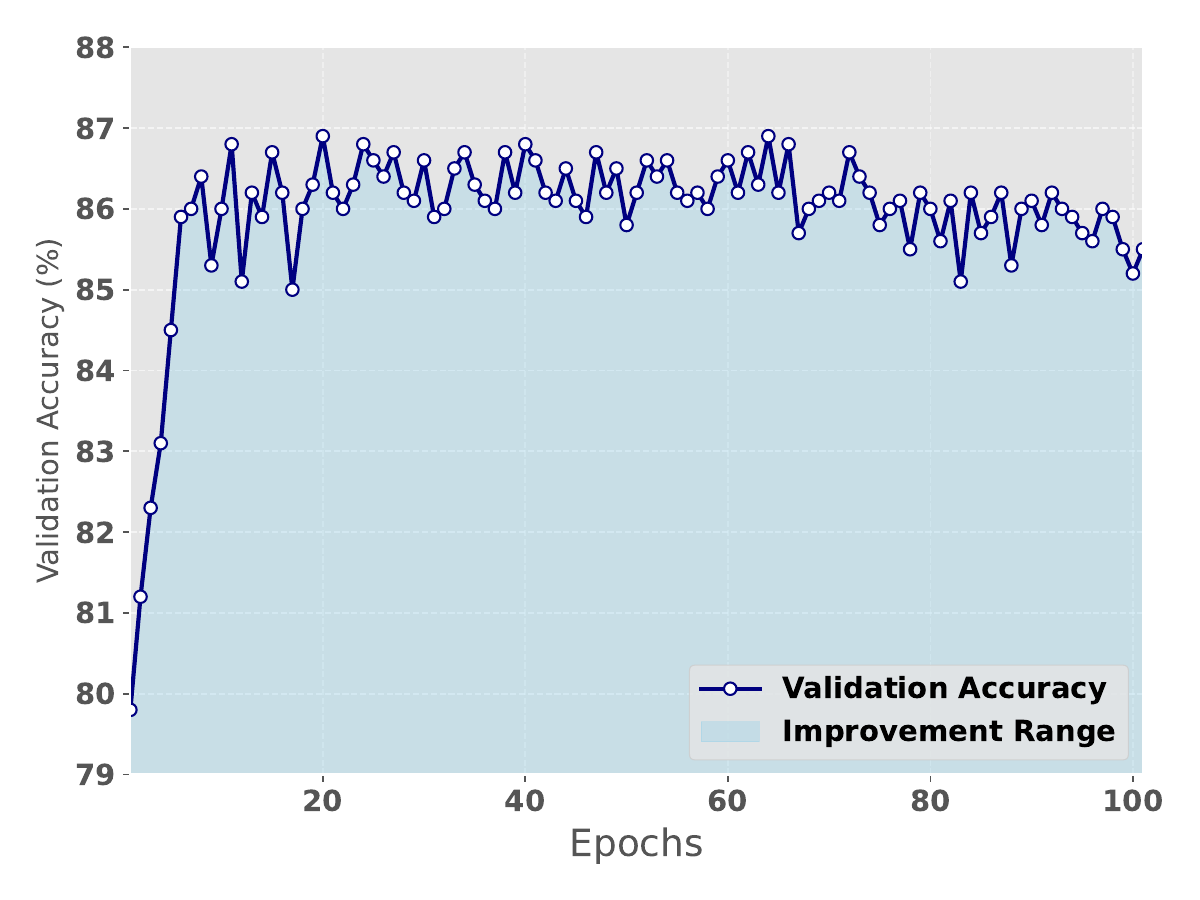}
        \caption*{(a)}  
        \label{fig:validation_accuracy}
    \end{subfigure}%
    \begin{subfigure}[b]{0.50\columnwidth}
        \centering
        \includegraphics[width=\linewidth]{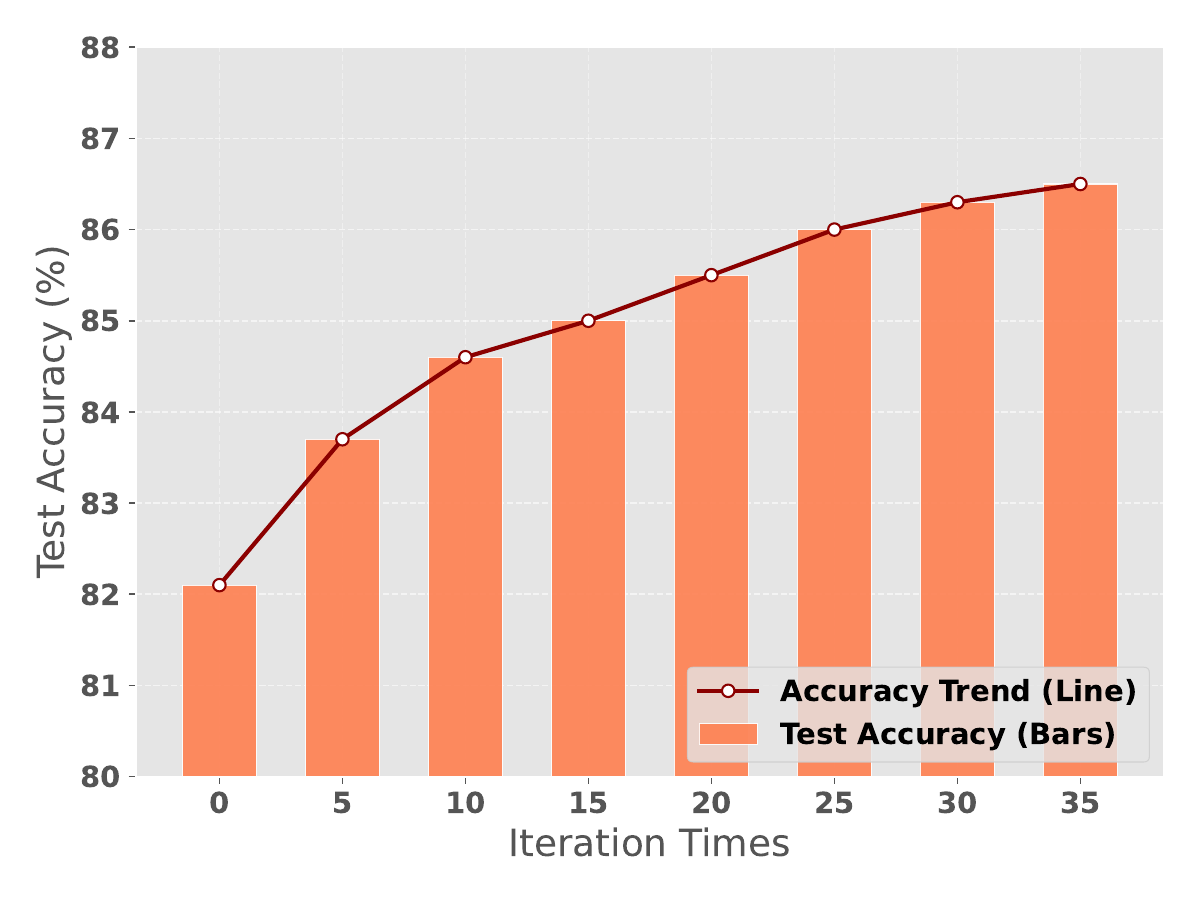}
        \caption*{(b)}  
        \label{fig:iterations_impact}
    \end{subfigure}
    \caption{Analysis of training dynamics: (a) Validation accuracy across epochs shows the learning behavior, and (b) inner-loop iterations highlight their influence on few-shot performance.}
    \label{fig:training_dynamics}
\end{figure}

We first analyzed the variation in validation accuracy during training, as shown in Figure~\ref{fig:training_dynamics}(a). A rapid improvement in early epochs highlights the model's efficient learning capability, while a slight decline beyond 80 epochs suggests overfitting. This observation underscores the importance of early stopping to balance computational efficiency and performance. Additionally, Figure~\ref{fig:training_dynamics}(b) illustrates the impact of inner-loop iterations, where accuracy peaks at 35 iterations, demonstrating the effectiveness of inner-loop optimization in enhancing generalization while maintaining computational efficiency.

\begin{figure}[ht]
    \centering
    \begin{subfigure}[b]{0.50\columnwidth}
        \centering
        \includegraphics[width=\linewidth]{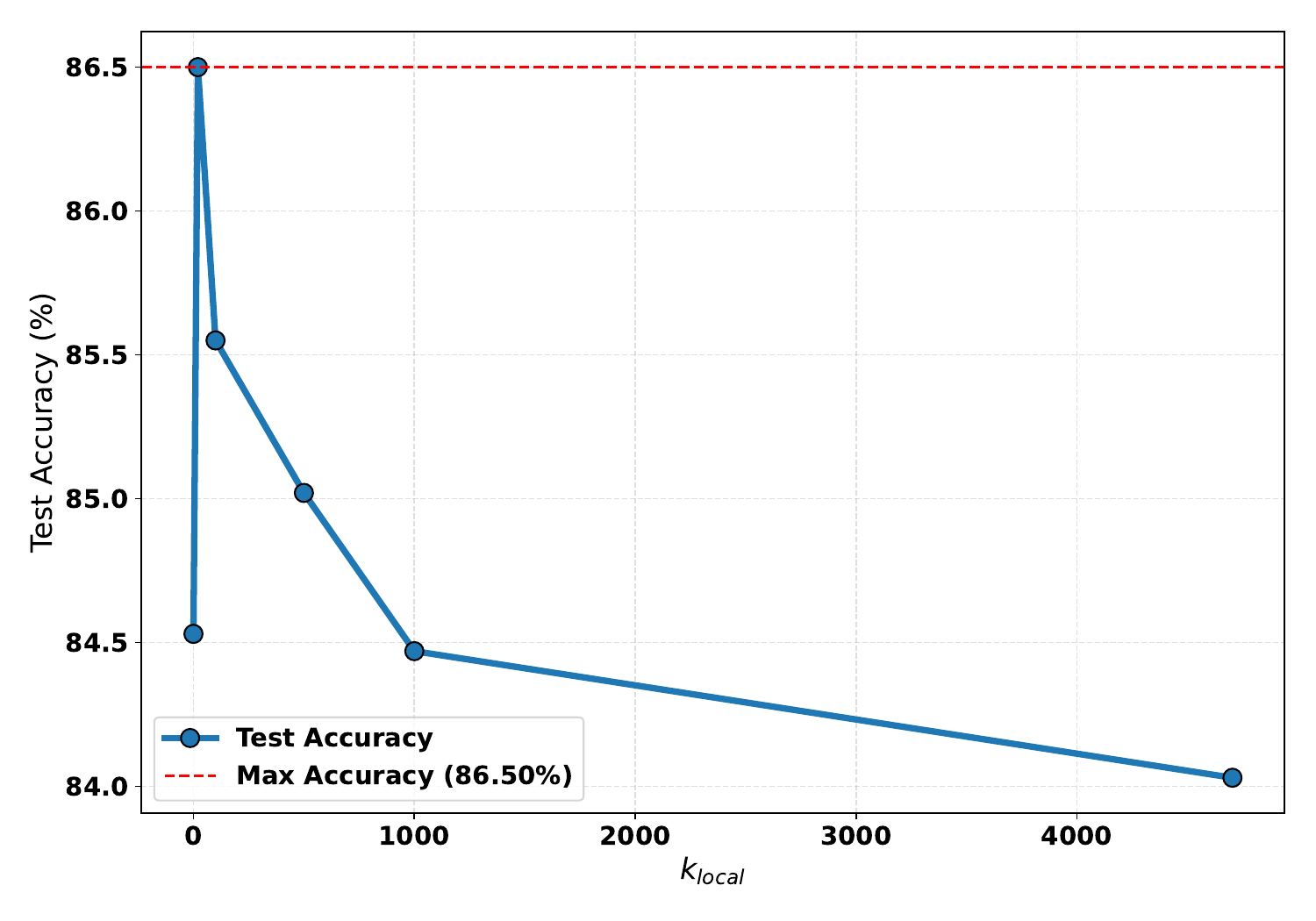}
        \caption*{(a)}  
        \label{fig:impact_k_local}
    \end{subfigure}%
    \begin{subfigure}[b]{0.50\columnwidth}
        \centering
        \includegraphics[width=\linewidth]{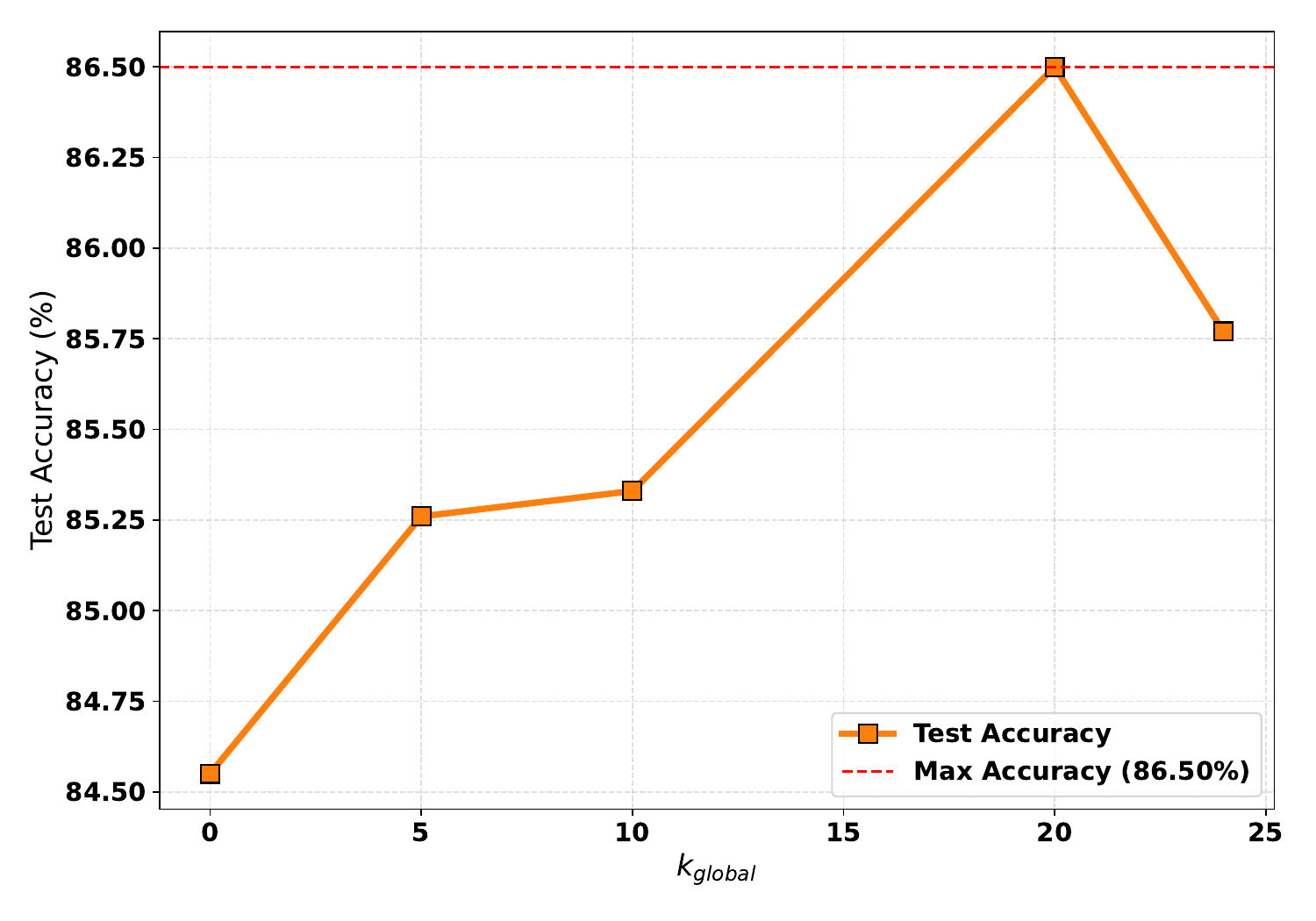}
        \caption*{(b)}  
        \label{fig:impact_k_global}
    \end{subfigure}
   \caption{Effect of $k_{\text{local}}$ (a) and $k_{\text{global}}$ (b) on test accuracy for the mini-ImageNet 5-way 5-shot task. (a) Number of local tokens retained for fine-grained interactions. (b) Number of global tokens for cross-cluster attention to maintain global semantics.}
    \label{fig:globa_local_tokens}
\end{figure}

Next, we evaluated the effect of \(k_{\text{local}}\) and \(k_{\text{global}}\) on classification accuracy, as depicted in Figures~\ref{fig:globa_local_tokens}(a) and~\ref{fig:globa_local_tokens}(b). These parameters regulate the token refinement process, with \(k_{\text{local}}\) focusing on fine-grained interactions within clusters and \(k_{\text{global}}\) emphasizing broader global relationships across clusters. For \(k_{\text{local}}\), fixing \(k_{\text{global}} = 20\) revealed that accuracy peaked at 86.50\% for \(k_{\text{local}} = 20\). This value balances meaningful feature retention and noise minimization. Larger values diluted refinement quality by including irrelevant tokens, while smaller values restricted the model’s ability to capture localized interactions. Similarly, fixing \(k_{\text{local}} = 20\), the highest accuracy (86.50\%) was achieved with \(k_{\text{global}} = 20\), suggesting an optimal balance between meaningful global connections and redundancy avoidance.

These findings highlight the critical role of \(k_{\text{local}}\), \(k_{\text{global}}\), class separation, clustering, uncertainty estimation, and regularization in optimizing our framework. Striking the right balance among these parameters is essential for achieving superior generalization and robust performance in few-shot learning tasks.

\subsection{Ablation Study}
To assess the contribution of each component in BATR-FST, we performed ablation experiments on the mini-ImageNet 5-way 1-shot and 5-shot tasks, as presented in Table \ref{tab:ablation_study}. The baseline (\textbf{I}) consists of pre-training alone, achieving 61.20\% for 1-shot and 73.80\%  for 5-shot, highlighting the limitations of pre-training for few-shot learning. Adding meta-finetuning (\textbf{II}) significantly improves performance to 68.35\% for 1-shot and 82.42\% for 5-shot, demonstrating the importance of task-specific adaptation. Incorporating the Bi-Level Attention Mechanism (\textbf{III}) further boosts accuracy to 69.90\%  for 1-shot and 85.10\%  for 5-shot by refining token representations and capturing local and global dependencies. The Graph Token Propagation (GTP) module (\textbf{IV}) enhances information flow between tokens, improving performance to 70.20\% for 1-shot and 86.10\% for 5-shot. Finally, adding the Uncertainty-Aware Token Weighting module (\textbf{V}) results in the highest accuracy, achieving 70.36\%  for 1-shot and 86.50\% for 5-shot by prioritizing reliable tokens. The ablation study highlights the complementary contributions of each component, with the complete configuration (\textbf{V}) achieving superior performance, validating the effectiveness of our proposed framework.

\begin{table}[ht!]
\centering
\caption{Ablation study results on the mini-ImageNet 5-way classification task. ``-'' denotes without the component, and ``\checkmark'' denotes with the component.}
\label{tab:ablation_study}
\resizebox{\columnwidth}{!}{%
\begin{tabular}{@{}lcccc|cc@{}}
\toprule
\textbf{Type} & 
\makecell{\textbf{Meta-}\\\textbf{Finetuning}} & 
\makecell{\textbf{Bi-Level}\\\textbf{Attention}} & 
\makecell{\textbf{Graph Token}\\\textbf{Propagation}} & 
\makecell{\textbf{Uncertainty-Aware}\\\textbf{Weighting}} & 
\makecell{\textbf{1-Shot}\\\textbf{Accuracy (\%)}} & 
\makecell{\textbf{5-Shot}\\\textbf{Accuracy (\%)}} \\ 
\midrule
(I) Baseline   & -  & -  & -  & -  & 61.20 & 73.80 \\
(II) Meta-Finetuning Only   & \checkmark  & -  & -  & -  & 68.35 & 82.42 \\
(III) With Bi-Level Attention   & \checkmark  & \checkmark  & -  & -  & 69.90 & 85.10 \\
(IV) With Graph Token Propagation   & \checkmark  & \checkmark  & \checkmark  & -  & 70.20 & 86.10 \\
(V) Full Configuration (Ours)   & \checkmark  & \checkmark  & \checkmark  & \checkmark  & \textbf{70.36} & \textbf{86.50} \\ 
\bottomrule
\end{tabular}%
}
\end{table}


\section{Conclusion}
This paper introduces BATR-FST, an innovative Bi-Level Adaptive Token Refinement method for Few-Shot Transformers. In the pre-training phase, we utilize Masked Image Modeling to extract resilient patch-level features, connecting self-supervised learning with few-shot contexts. During the meta-finetuning phase, our Bi-Level Adaptive Token Refinement mechanism dynamically enhances token representations at local and global levels, improving flexibility to various few-shot tasks. The key components include Uncertainty-Aware Token Weighting for assessing token reliability, a Bi-Level Attention mechanism to balance local clustering with global context, Graph Token Propagation for understanding dependencies between support and query instances, and a Class Separation Penalty to enhance intra-class compactness while maintaining inter-class distinctiveness. The attained results on three benchmark datasets and additional experimental results discussed in this study illustrate the effectiveness and competitiveness of our proposed BATR-FST method.

\section*{Acknowledgment}

This work was supported by the National Natural Science Foundation of China (Grant No. 72061147004).

\bibliographystyle{IEEEtran}
\bibliography{BATR}

\end{document}